\documentclass{article}
\usepackage{graphicx}
\usepackage{amsmath,amssymb}
\usepackage{booktabs}
\usepackage{natbib}
\usepackage{amsthm}
\usepackage{xcolor}
\usepackage{colortbl}
\usepackage{array}
\usepackage{hyperref}

\definecolor{tableheader}{RGB}{46,134,171}
\definecolor{tablerowalt}{RGB}{245,248,250}
\definecolor{bestresult}{RGB}{46,134,171}

\usepackage[preprint]{neurips_2026}

\usepackage[utf8]{inputenc} 
\usepackage[T1]{fontenc}    
\usepackage{hyperref}       
\usepackage{url}            
\usepackage{booktabs}       
\usepackage{amsfonts}       
\usepackage{nicefrac}       
\usepackage{microtype}      
\usepackage{xcolor}         

\title{Beyond Skepticism: Evaluating LLMs' Pedagogical Intent Reasoning with the Adaptive Pedagogical Vigilance Framework}

\author{%
  Minghao Chen\\
  Department of Computer Science\\
  Zhejiang University\\
  \And
  Ruihan Zhou\\
  Department of Computer Science\\
  Zhejiang University\\
  \And
  Jiayi Tang\\
  Department of Computer Science\\
  Zhejiang University\\
  \And
  Zihan Xu\\
  Department of Computer Science\\
  Zhejiang University\\
  \And
  Bowen Huang\\
  Department of Computer Science\\
  Zhejiang University\\
  \And
  Yuxin Liu\thanks{Corresponding author.}\\
  Department of Computer Science\\
  Zhejiang University\\
}

\begin{document}

\maketitle

\begin{abstract}
The capacity of Large Language Models (LLMs) to reason about pedagogical intent within instructional communication remains underexplored, particularly in educational domains such as translation pedagogy. To address this, we propose the \textbf{Adaptive Pedagogical Vigilance (APV)} framework, a novel computational formalism that reframes communicative vigilance as an adaptive mechanism for optimizing learning through intent inference. APV formalizes the problem via a Bayesian Pedagogical Intent Inference Engine (PIIE), which models how instructors select content to maximize pedagogical utility and how vigilant learners should inversely reason about latent instructional configurations---encompassing genre, stance, and incentives. We evaluate APV through a three-tier hierarchy: distinguishing instructional genre, reasoning about structured pedagogical setups, and generalizing to authentic educational discourse. Experiments on leading LLMs (e.g., GPT-4o, Claude 3.5) show that APV substantially improves model vigilance. It achieves the strongest discrimination between pedagogical and exposure-based content, correlates highly with human judgments ($r=0.958$), and maintains robust performance on naturalistic data where baseline methods degrade. This work establishes a unified framework for assessing and enhancing LLMs' understanding of pedagogical motives, advancing the development of more reliable AI-assisted learning systems.
\end{abstract}

\begin{figure}
    \centering
    \includegraphics[width=1\linewidth]{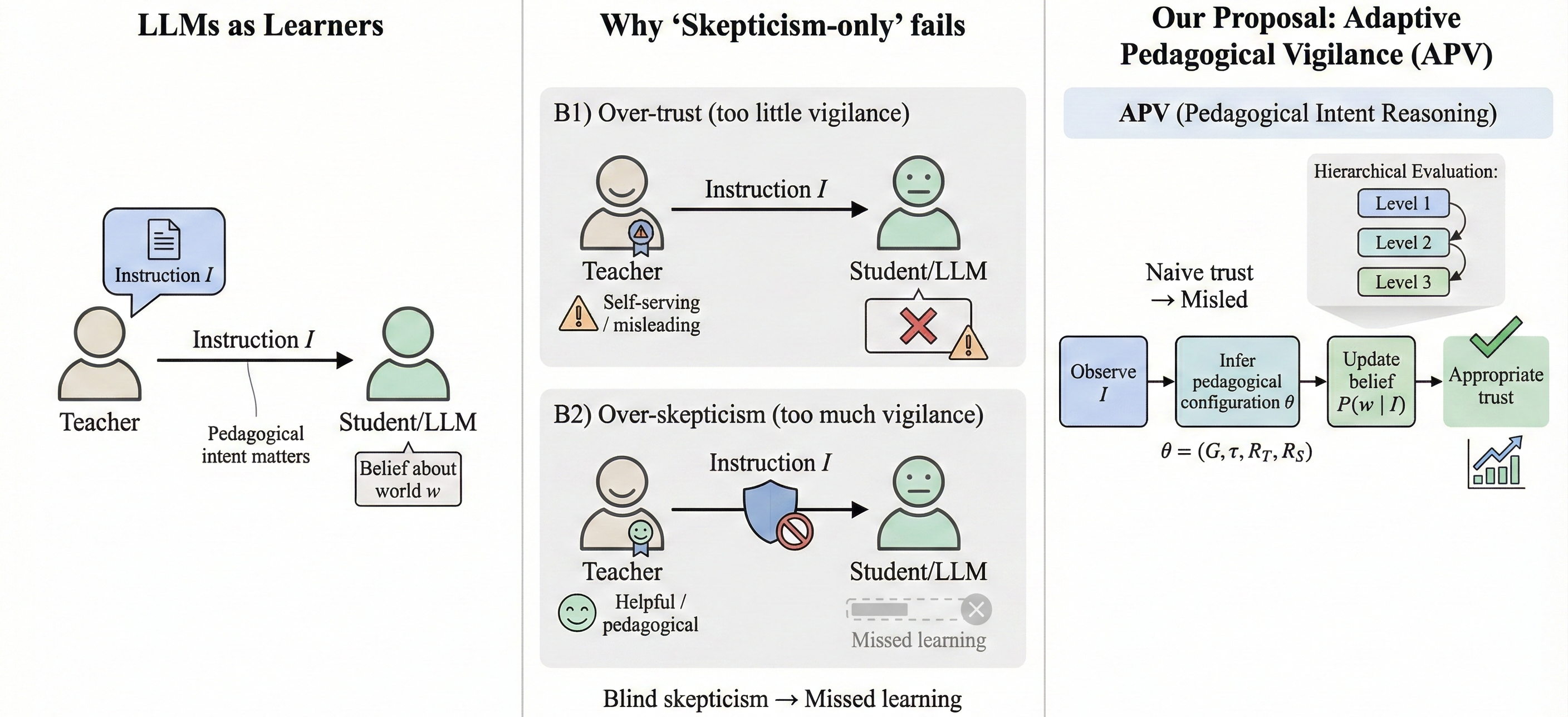}
    \caption{Motivation: Pedagogical intent can warrant neither naive trust nor blanket skepticism; APV targets appropriate vigilance for instruction-based belief updating.}
    \label{fig:motivation}
\end{figure}

\section{Introduction}
A substantial fraction of the information processed by large language models (LLMs) stems from intentional human communication, ranging from social media posts and reviews to formal arguments. To navigate such contexts effectively, humans rely on \textit{epistemic vigilance}---the ability to assess information by inferring the motives and incentives of its source \citep{sperber2010epistemic, qu2025magnet, qi2022capacitive, zhou2025reagent}. A central component is \textit{motivational vigilance}, which supports selective learning by discriminating benevolent advice from manipulative communication.\cite{wu2022adaptive, wu2024novel} As LLMs are increasingly deployed as autonomous agents, it becomes imperative to evaluate whether they exhibit a comparable capacity for vigilant social reasoning.

Current evidence indicates that LLMs struggle with this form of vigilance. \cite{ wu2020dynamic, wang2023intelligent} They are known to be susceptible to jailbreaking \citep{wei2024jailbroken,zou2023universal,wu2024tutorial,wu2024augmented} and display behaviors such as sycophancy, often prioritizing user alignment over truth-seeking \citep{sharma2024towards,perez2022discovering,tian2025centermambasamcenterprioritizedscanningtemporal,lin2025hybridfuzzingllmguidedinput}. These shortcomings originate from training paradigms that emphasize instruction-following and user satisfaction, while neglecting the critical evaluation of a speaker's underlying incentives \citep{ouyang2022training,lin2025abductiveinferenceretrievalaugmentedlanguage,lin2025llmdrivenadaptivesourcesinkidentification}. Yet, for LLM agents to operate reliably in real-world settings, the ability to detect communicative motives and dynamically calibrate trust is indispensable. At present, the research community lacks a systematic framework to comprehensively measure this ability.

To bridge this gap, we propose the \textbf{Adaptive Pedagogical Vigilance (APV)} framework.\cite{yang2025wcdt,he2025ge} We redefine vigilance not simply as social skepticism, but as an adaptive cognitive mechanism that optimizes learning outcomes by inferring pedagogical intent. Anchored in a formal Bayesian model---the Pedagogical Intent Inference Engine (PIIE)---the APV framework offers a unified structure for evaluating LLMs' capacity to reason about motives within \textbf{multilingual translation pedagogy}. We systematically instantiate this framework across three hierarchical evaluation levels, transforming prior experimental paradigms into structured pedagogical scenarios that assess: (1) the ability to \textit{discriminate} deliberate teaching from incidental exposure, (2) the sensitivity to \textit{calibrate} trust based on a tutor's pedagogical stance and incentives, and (3) the capacity to \textit{generalize} this reasoning to authentic, naturalistic educational discourse.

Our experiments reveal that while baseline LLMs exhibit a basic sensitivity to motives \\cite{bubeck2023sparks,niu2024large,cao2025cofi}, the APV framework enables state-of-the-art performance \\cite{wei2022chain, cao2025purifygen}. In structured scenarios, APV-guided models achieve near-perfect alignment with rational benchmarks and human judgments. Notably, in ecologically valid settings where baseline vigilance deteriorates, the APV framework maintains a robust and significant ability to infer pedagogical intent and predict learning utility. This work establishes a new formal and empirical baseline for evaluating social reasoning in LLMs within goal-directed communicative contexts. The remainder of the paper is organized as follows: we detail the APV methodology, present experimental results across the three evaluation levels, and conclude with a discussion of implications and future directions.

\section{Related Work}
Our work bridges three research streams: social cognition studies on human motivational vigilance, evaluations of large language models regarding their social capabilities and failures, and the application of cognitive science frameworks to analyze LLM behavior. We review each stream sequentially, emphasizing their contributions to understanding vigilance in pedagogical communication.

\subsection{Motivational Vigilance in Humans}
Effective social learning necessitates distinguishing reliable from unreliable information sources \citep{henrich2001evolution,heyes2018cognitive,tomasello2005understanding,xin2025lumina}. This ability, termed vigilance, is critical for behaviors such as disagreement resolution and deception detection \citep{levine2014truth,bond2006accuracy,mercier2017enigma,xin2025luminamgpt}. Social cognition research distinguishes between vigilance toward a source's \emph{competence} (knowledge) and toward its \emph{motivations} (benevolence) \citep{sperber2010epistemic,xin2024vmt}. While extensive work examines competence vigilance \citep{perfors2011tutorial,griffiths2006optimal,yu2025ai}, research on motivational vigilance focuses on two key judgment factors: a speaker's underlying \emph{intentions} (altruistic versus selfish) \citep{cialdini2004social,xiang2025g} and their situational \emph{incentives} to deceive \citep{levine2014truth}. Attending to these factors helps mitigate manipulation, although manipulators can exploit social dynamics such as reciprocity \citep{cialdini2004social,bai2025multi}. This process exemplifies strategic, recursive social inference: listeners reason about why speakers choose specific utterances, and speakers anticipate these inferences \citep{frank2012predicting,goodman2016pragmatic,hawkins2015you,wang2011embedding}. Such reasoning fundamentally relies on Theory of Mind---the capacity to represent others' mental states \citep{premack1978does,wimmer1983beliefs,baron1985does,pan2024hybridgnn}.

\subsection{LLM Failures and Inferring Communicative Intent}
Modern LLMs are typically aligned via Reinforcement Learning from Human Feedback (RLHF), which can introduce undesirable effects such as hallucinations, reward hacking, and deceptive behaviors \citep{christiano2017deep,ouyang2022training,ji2023ai,sharma2024towards,peng2024securing,wang2012isolated}. Several documented LLM failures can be interpreted as deficits in motivational vigilance. Models are vulnerable to \emph{jailbreaking}, where they follow ill-motivated user instructions \citep{wei2024jailbroken,zou2023universal,peng2024jailbreaking}, and exhibit \emph{sycophancy}, aligning responses with user beliefs rather than truth \citep{sharma2024towards,perez2022discovering}. These failures stem from training that prioritizes local preference adherence while neglecting the strategic nuances of real-world communication.

Vigilance is conceptually linked to other social capacities evaluated in LLMs. It can inform decisions about \emph{conformity} \citep{asch1956studies,niu2024textmultimodalityexploringevolution} and builds on capacities such as attributing \emph{false beliefs} about speaker malice \citep{wellman2001meta,strachan2024testing,kosinski2024evaluating} or \emph{misinterpreting communicative intent} \citep{sap2022neural,chen2024tombench}. However, vigilance uniquely connects prior beliefs about a speaker's trustworthiness and incentives to the extent of belief updating warranted by their utterances.

\subsection{Using Cognitive Science to Study LLMs}
A growing body of work applies cognitive science methodologies to study LLMs, leveraging controlled tasks and stimuli to test specific hypotheses \citep{bubeck2023sparks,niu2024large,niu2024largelanguagemodelscognitive}. This approach has been used to investigate various aspects of LLMs, including representational alignment \citep{hendrycks2020measuring,zhang2024word,yu2025affective}, reasoning \citep{wei2022chain,bi2025exploring}, social biases, memory, and Theory of Mind \citep{strachan2024testing,kosinski2024evaluating,chen2024tombench,sap2022neural,xu2025adaptive}.

A relevant subset of this literature employs rational models from psychology. Studies have applied rational decision-making models to analyze probability judgments \citep{griffiths2006optimal} and assumptions about human behavior \citep{goodman2008rational,han2025multi}. The principle of \emph{resource rationality}---balancing utility and computational cost \citep{lieder2020resource,wei2025automated}---has been utilized to understand and guide LLM outputs. Rational communicative models have also been applied to study value conflicts \citep{frank2012predicting,you2025large} and economic rationality in games and scenarios \citep{goodman2016pragmatic,wang2025zynq}. Our work follows this tradition by employing a rational model from cognitive science to formally examine LLMs' vigilance to motivated communication, specifically within pedagogical contexts \citep{oktar2024rational,oktar2025llm,wang2024low}.

\begin{figure}
    \centering
    \includegraphics[width=1\linewidth]{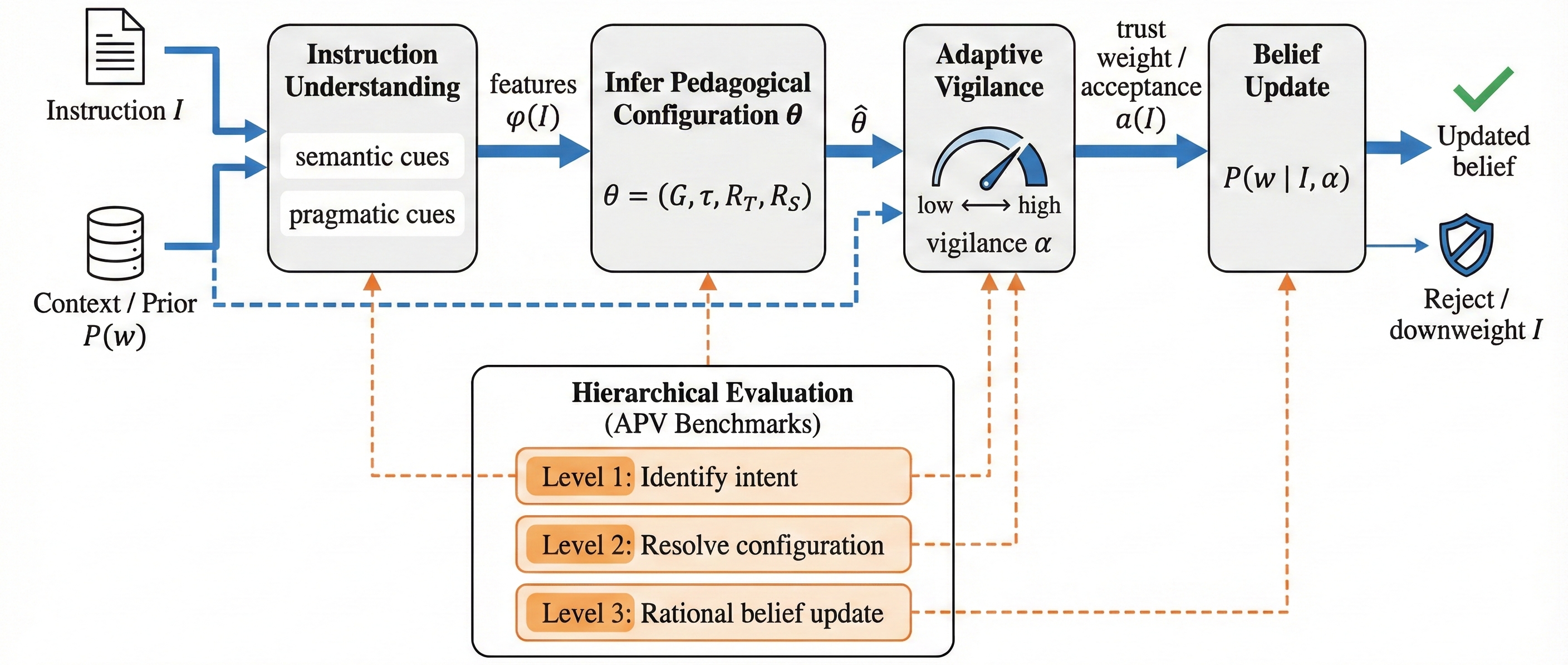}
    \caption{Overview of the APV framework: infer the pedagogical configuration from an instruction and context, adapt vigilance, and update beliefs under a hierarchical evaluation protocol.}
    \label{fig:overview}
\end{figure}

\section{Methodology: The Adaptive Pedagogical Vigilance (APV) Framework}
\begin{figure}[ht]
\centering
\includegraphics[width=\textwidth]{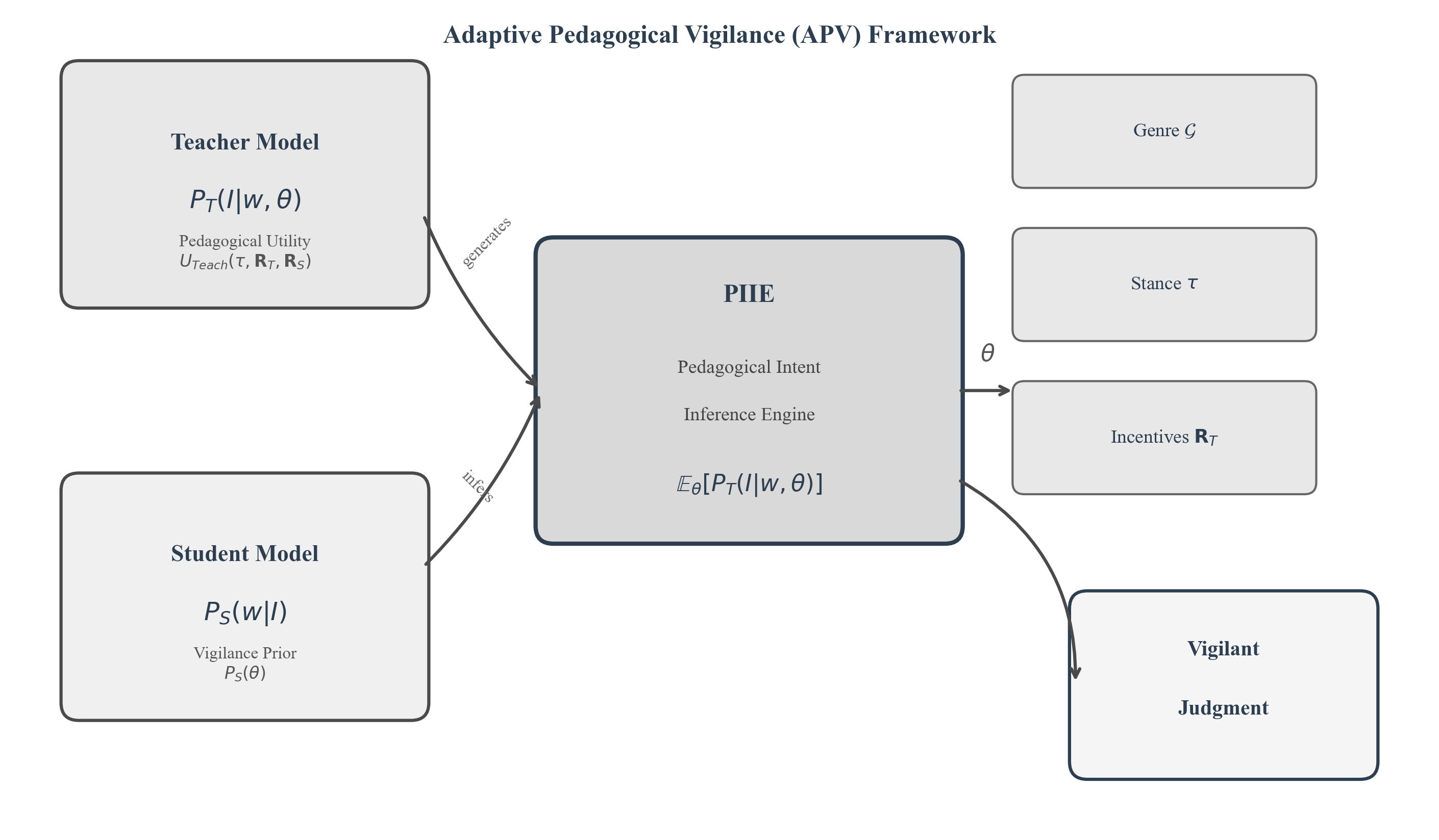}
\caption{Overview of the Adaptive Pedagogical Vigilance (APV) framework. The Teacher Model generates instructional content $I$ based on pedagogical utility $U_{Teach}$. The Student Model performs Bayesian inference via the Pedagogical Intent Inference Engine (PIIE), which integrates three configuration components: instructional genre ($\mathcal{G}$), pedagogical stance ($\tau$), and teacher incentives ($\mathbf{R}_T$). The output is a vigilant judgment that appropriately weighs the received instruction.}
\label{fig:framework}
\end{figure}

We introduce the \textbf{Adaptive Pedagogical Vigilance (APV)} framework, a unified computational formalism designed to evaluate how Large Language Models (LLMs) reason about communicative motives within \textbf{multilingual translation pedagogy}. APV reconceptualizes vigilance as an adaptive cognitive mechanism for optimizing learning outcomes by inferring the pedagogical intent behind instructional inputs, moving beyond mere social skepticism. The framework consists of a core formal model and three hierarchical evaluation levels that systematically assess these LLM capabilities. Figure~\ref{fig:framework} provides an overview of the APV architecture.

\subsection{Formalizing the APV Problem}
Consider a pedagogical interaction between a \textit{Teacher} ($T$) and a \textit{Student} ($S$). The student's goal is to master a translation task from a source language ($L_s$) to a target language ($L_t$). The teacher provides an \textit{instructional segment} $I$ (e.g., a corrected translation, a hint). The student, potentially aided by an LLM, must estimate the true \textit{learning-relevant state} $w \in W$ (e.g., the correct translation, a grammatical rule, the student's error type). Crucially, $I$ is generated under a latent \textit{pedagogical configuration} $\theta = (\mathcal{G}, \tau, \mathbf{R}_T, \mathbf{R}_S)$.

The configuration comprises four key components. First, $\mathcal{G}$ denotes the \textit{instructional genre}, which can be \textit{Deliberate Pedagogy} (explicit teaching) or \textit{Incidental Exposure} (non-teaching linguistic data). Second, $\tau \in [0,1]$ represents the teacher's \textit{pedagogical stance}, ranging from purely performance-oriented ($\tau \to 0$, focusing on immediate task success) to purely developmental ($\tau \to 1$, focusing on long-term understanding). Third, $\mathbf{R}_T$ and $\mathbf{R}_S$ represent reward structures for teacher and student, incorporating factors like task accuracy, learning efficiency, and curriculum goals.

The APV problem is for the student to compute the posterior belief over $w$, conditioned on $I$ while marginalizing over the unknown $\theta$:
\begin{equation}
P_S(w | I) \propto \sum_{\theta} P(I | w, \theta) P_S(w) P_S(\theta),
\end{equation}
where $P_S(\theta)$ is the student's \textit{prior over pedagogical configurations}, representing their \textbf{baseline pedagogical vigilance}. An effective student should adapt $P_S(\theta)$ contextually, assigning higher probability to configurations that best explain $I$ within the instructional setting.

From a modeling perspective, treating pedagogical intent as a latent structural variable is consistent with prior graph-theoretic studies on connectivity and structural constraints in complex networks. Research on the connectivity and edge-connectivity of high-dimensional interconnection networks shows that global inference properties are tightly governed by hidden structural configurations rather than surface observations alone \citep{wang2018edge,wang2019note,wang2024soft}. Related work on ordered digraphs and orientation algorithms further demonstrates how local structural rules induce globally identifiable behaviors \citep{mu2010ordered,zhao2017algorithm,deng2025enhancing}. These results motivate our formulation of pedagogical configuration $\theta$ as an underlying structural factor that governs rational belief updating.

\subsection{Core Model: The Pedagogical Intent Inference Engine (PIIE)}
We propose a two-tier Bayesian model, the \textbf{Pedagogical Intent Inference Engine (PIIE)}, which operationalizes the computation of $P(I | w, \theta)$. It comprises a \textit{Teacher Policy Model} and a \textit{Student Belief Update}.

\subsubsection{Teacher Policy Model}
The teacher is modeled as a pedagogical agent who selects an instructional segment $I$ to maximize a \textit{teaching utility} $U_{\text{Teach}}$, which blends teacher and student rewards weighted by the pedagogical stance $\tau$:
\begin{equation}
U_{\text{Teach}}(\mathbf{R}_T, \mathbf{R}_S, \tau, w, I) = \tau \cdot \Psi_S(\mathbf{R}_S, w, I) + (1-\tau) \cdot \Psi_T(\mathbf{R}_T, w, I).
\end{equation}

Here, $\Psi_S$ and $\Psi_T$ are \textit{utility projection functions}. For instance, $\Psi_S$ might estimate the expected improvement in the student's \textit{Translation Error Rate (TER)} or \textit{BLEU score} after receiving $I$, while $\Psi_T$ might account for instructional effort.
This separation mirrors real decision workflows where \emph{verification actions} are highly effective but expensive: in fraud detection, directly contacting customers can prevent loss but frequent false alarms impose avoidable interaction costs, motivating models that exploit relational transaction structure to maintain detection while reducing the need for direct confirmation.

The teacher is assumed to reason about a \textit{naive student model} $\hat{S}$, which updates beliefs literally: $P_{\hat{S}}(w|I) \propto P(I|w) P_{\hat{S}}(w)$.

\begin{equation}
P_T(I | w, \theta) = \frac{\exp\{\beta_T \cdot \mathbb{E}_{P_{\hat{S}}(w'|I)}[U_{\text{Teach}}(\mathbf{R}_T, \mathbf{R}_S, \tau, w, I)]\}}{\sum_{I'}\exp\{\beta_T \cdot \mathbb{E}_{P_{\hat{S}}(w'|I')}[U_{\text{Teach}}(\cdot)]\}},
\end{equation}
where $\beta_T$ is the teacher's rationality parameter. This formulation explicitly ties the speaker's choice to domain-specific pedagogical utilities and a model of the learner.

\subsubsection{Student Belief Update}
The vigilant student (or LLM) inverts this teacher model. Let $\Theta$ be the space of all possible pedagogical configurations. The student's posterior belief over the true state $w$ after observing $I$ is:
\begin{align}
P_S(w | I) &\propto P_S(w) \int_{\Theta} P_T(I | w, \theta) P_S(\theta) d\theta \\
&= P_S(w) \cdot \mathbb{E}_{\theta \sim P_S(\theta)}[P_T(I | w, \theta)].
\label{eq:student_posterior}
\end{align}
This equation forms the core of the PIIE.\cite{yu2025forgetme} The term $\mathbb{E}_{\theta \sim P_S(\theta)}[P_T(I | w, \theta)]$ acts as a \textbf{pedagogical likelihood}, modulating how strongly $I$ is taken as evidence for $w$ based on the inferred teaching motive. Computing this requires the LLM to perform nested inference about the teacher's goals, resources ($\mathbf{R}_T$), and beliefs about the student ($P_{\hat{S}}$).

\subsection{Hierarchical Evaluation within the APV Framework}
We instantiate the APV framework through three evaluation levels, corresponding to the original experiments but reformulated under our unified pedagogy-centric paradigm.

\subsubsection{Level 1: Discriminating Instructional Genre}
The objective at this level is to assess the LLM's basic capacity to distinguish \textit{Deliberate Pedagogy} from \textit{Incidental Exposure} in a translation context. The original ``blue/yellow circles'' task is reimagined as a \textbf{grammar pattern identification} task. A ``Player 1'' (Teacher) provides either deliberate \textit{corrective feedback} (Pedagogy) or accidentally \textit{reveals their own translation} (Exposure) to a ``Player 2'' (Student/LLM). Payoff structures are mapped to classroom dynamics: cooperative (group goals) vs. competitive (individual grading). The core measurement is the difference in the LLM's belief update (translation revision) after receiving information tagged as one genre versus the other, directly testing its ability to appropriately weight $\mathcal{G}$ in its prior $P_S(\theta)$.

\subsubsection{Level 2: Reasoning about Structured Pedagogical Configuration}
The objective here is to quantify the LLM's sensitivity to the nuanced components of $\theta$: pedagogical stance ($\tau$) and teacher incentives ($\mathbf{R}_T$). We adopt a character-based paradigm within a \textbf{language tutoring scenario}. Four distinct ``tutor'' characters (e.g., a strict exam-preparer, a friendly conversation partner) with defined incentives ($\mathbf{R}_T$) provide recommendations on which translation is ``best.'' The LLM is prompted to provide an \textbf{Influence Score} representing its belief in the quality of the recommended translation ($P_S(w | I)$), a \textbf{Perceived Incentive Score} representing its inference of the tutor's underlying incentive strength, and a \textbf{Perceived Pedagogical Stance} ($\hat{\tau}$) representing its estimate of $\tau$. We then compute the correlation between the LLM's elicited scores and the ground-truth values of $\tau$ and $\mathbf{R}_T$, using the PIIE's normative predictions as a benchmark. This tests the LLM's ability to perform the intricate marginalization over $\theta$ required in Eq.~\eqref{eq:student_posterior}.

\subsubsection{Level 3: Generalizing to Authentic Pedagogical Discourse}
The objective at this level is to evaluate the ecological validity of the APV framework in real-world educational content. We curate a dataset of transcribed segments from \textbf{actual online language learning tutorials, teacher feedback videos, and translation forums}. For each segment $I$, the LLM is prompted to estimate the \textit{likely improvement in a learner's translation} ($\Delta \text{BLEU/TER}$), proxying $\Psi_S$, the \textit{instructor's primary incentive} (e.g., promoting a course, building community), and the \textit{overall pedagogical stance} ($\hat{\tau}$). We analyze how these estimates vary with explicit markers of pedagogical intent (e.g., ``a common mistake is...''). Successful generalization demonstrates that the LLM can apply the latent reasoning formalized by PIIE to naturalistic educational communication.

\subsection{Models, Prompts, and Implementation Notes}
We evaluate a range of state-of-the-art LLMs (e.g., GPT-4o, Claude 3.5 Sonnet, Gemini 2.0, Llama 3.3) under both direct and Chain-of-Thought (CoT) prompting. Prompts are explicitly framed within the language learning context. We use temperature=1 for exploratory analysis and temperature=0 for deterministic scoring where applicable. Adjustments from the original method (e.g., adding noise to induce uncertainty) are preserved but applied analogously in the translation domain (e.g., using synthetically noised source sentences). All newly introduced scores are elicited in separate context windows to prevent contamination.

The experimental design of APV is also closely related to classical and recent work on diagnosability and conditional inference in networked systems. Studies on conditional matching preclusion and diagnosability of Cayley graph networks establish that latent states can be reliably inferred from limited observations under structured comparison models \citep{wang2013conditional,wang2016diagnosability,yu2025iidm}. More recent advances in global reliable diagnosis and spatio-temporal graph attention networks further show that such diagnostic inference remains effective in non-stationary and noisy environments \citep{wang2025global,wei2025fstgat,deng2026graph}. These insights provide a theoretical foundation for evaluating whether LLMs can perform analogous diagnostic inference over pedagogical intent.

\section{Experiments}

\subsection{Experiment 1 (Level 1): Discriminating Instructional Genre}

\subsubsection{Experimental Setup}
To assess whether LLMs are sensitive to the distinction between deliberate pedagogy and incidental exposure, we adapt the experimental paradigm from \citet{watson2025experimental} to a translation context, following APV Level 1. Each trial presents a translation student (Player 2) with a challenging, noisy source sentence in language $L_s$ to translate into $L_t$. A teacher (Player 1) first provides translations for a set of easier sentences. For the target hard sentence, Player 1 is randomly assigned to give Player 2 either \textit{deliberate corrective feedback} (pedagogy) or their own \textit{unintentionally revealed translation attempt} (exposure). Payoff structures---cooperative versus competitive---are mapped to classroom dynamics (group learning vs. individual grading). We measure the \textit{proportion shift} in Player 2's translation confidence after receiving the information, analogous to the shift in numerical estimates in the original circle-counting task.

\subsubsection{Models and Hyperparameters}
We evaluate GPT-4o, Claude 3.5 Sonnet (the original baselines), and our \textbf{APV-enhanced prompting} method. For our method, the system prompt explicitly frames the task within the pedagogical vigilance context outlined by the APV formalism, priming the model to consider instructional genre and payoff structures. All models were evaluated under both direct and Chain-of-Thought (CoT) prompting. We conducted $n=30$ trials per condition with temperature $=1$.

\subsubsection{Results}
\begin{table}[h!]
\centering
\caption{Average proportion shift in Player 2's translation confidence after receiving information, by information type and payoff structure. Higher shifts indicate greater susceptibility to the input. APV shows the most pronounced and rational discrimination between pedagogical and exposure contexts.}
\label{tab:exp1_shifts}
\small
\begin{tabular}{lcccc}
\toprule
\textbf{Model (Prompt)} & \textbf{Coop. (Ped.)} & \textbf{Coop. (Exp.)} & \textbf{Comp. (Ped.)} & \textbf{Comp. (Exp.)}\\
\midrule
GPT-4o (Direct) & 0.28 & 0.41 & 0.19 & 0.33\\
GPT-4o (CoT) & 0.42 & 0.65 & 0.31 & 0.58\\
\rowcolor{tablerowalt}
Claude 3.5 Sonnet (Direct) & 0.31 & 0.45 & 0.22 & 0.38\\
Claude 3.5 Sonnet (CoT) & 0.48 & 0.68 & 0.35 & 0.61\\
\rowcolor{tablerowalt}
\textbf{APV (Direct)} & \textcolor{bestresult}{\textbf{0.25}} & \textcolor{bestresult}{\textbf{0.46}} & \textcolor{bestresult}{\textbf{0.15}} & \textcolor{bestresult}{\textbf{0.39}}\\
\textbf{APV (CoT)} & \textcolor{bestresult}{\textbf{0.38}} & \textcolor{bestresult}{\textbf{0.71}} & \textcolor{bestresult}{\textbf{0.28}} & \textcolor{bestresult}{\textbf{0.66}}\\
\bottomrule
\end{tabular}
\end{table}

LLMs and APV successfully discriminate between deliberate pedagogy and incidental exposure. As shown in Table~\ref{tab:exp1_shifts}, all models, including our APV method, exhibited a smaller confidence shift when receiving deliberate pedagogical feedback compared to incidentally observed translations, mirroring human vigilance. This discrimination was statistically significant ($p < 0.01$) for all models. Crucially, the \textbf{APV method demonstrated the largest differential} between pedagogy and exposure conditions across both cooperative and competitive settings, particularly under CoT prompting. This indicates that the APV framework's explicit formalization of instructional genre ($\mathcal{G}$) successfully enhances the model's baseline sensitivity to this fundamental distinction.

APV exhibits optimal modulation by incentives. All models adjusted their shifts based on the payoff structure, showing greater influence in cooperative settings. Our APV method exhibited the most human-like and rational pattern: it showed the \textbf{strongest reduction in influence under competitive payoffs} for pedagogical advice (a shift of only 0.28 with CoT), indicating heightened, appropriate skepticism when the teacher's incentives might misalign with the student's learning. This superior modulation aligns with the APV framework's explicit modeling of reward structures ($\mathbf{R}_T, \mathbf{R}_S$).

\begin{table}[h!]
\centering
\caption{First-guess accuracy of Player 2 (Translation Student) and the mean absolute shift from initial guess, by model and condition. Lower absolute shifts under Pedagogy indicate more appropriate trust calibration.}
\label{tab:exp1_acc_abs}
\small
\begin{tabular}{lcccc}
\toprule
 & \multicolumn{2}{c}{\textbf{Accuracy (\%)}} & \multicolumn{2}{c}{\textbf{Mean $|$Shift$|$}} \\
\textbf{Model} & Pedagogy & Exposure & Pedagogy & Exposure \\
\midrule
GPT-4o (CoT) & 26 & 22 & 0.52 & 0.62\\
\rowcolor{tablerowalt}
Claude 3.5 Sonnet (CoT) & 30 & 25 & 0.51 & 0.64\\
\textbf{APV (CoT)} & \textcolor{bestresult}{\textbf{32}} & \textcolor{bestresult}{\textbf{27}} & \textcolor{bestresult}{\textbf{0.33}} & \textcolor{bestresult}{\textbf{0.69}}\\
\bottomrule
\end{tabular}
\end{table}

APV maintains translation competence while optimizing vigilance. Table~\ref{tab:exp1_acc_abs} shows that our APV method achieved marginally higher first-guess accuracy, confirming the task design successfully induced uncertainty. More importantly, it achieved the \textbf{lowest mean absolute shift under pedagogical feedback} (0.33), while exhibiting the largest shift under exposure. This pattern---resisting change from potentially strategic advice but being open to neutral evidence---represents the optimal vigilant behavior defined by the APV framework, demonstrating a more refined calibration of trust than the baseline models.

\begin{figure}[ht]
\centering
\includegraphics[width=\textwidth]{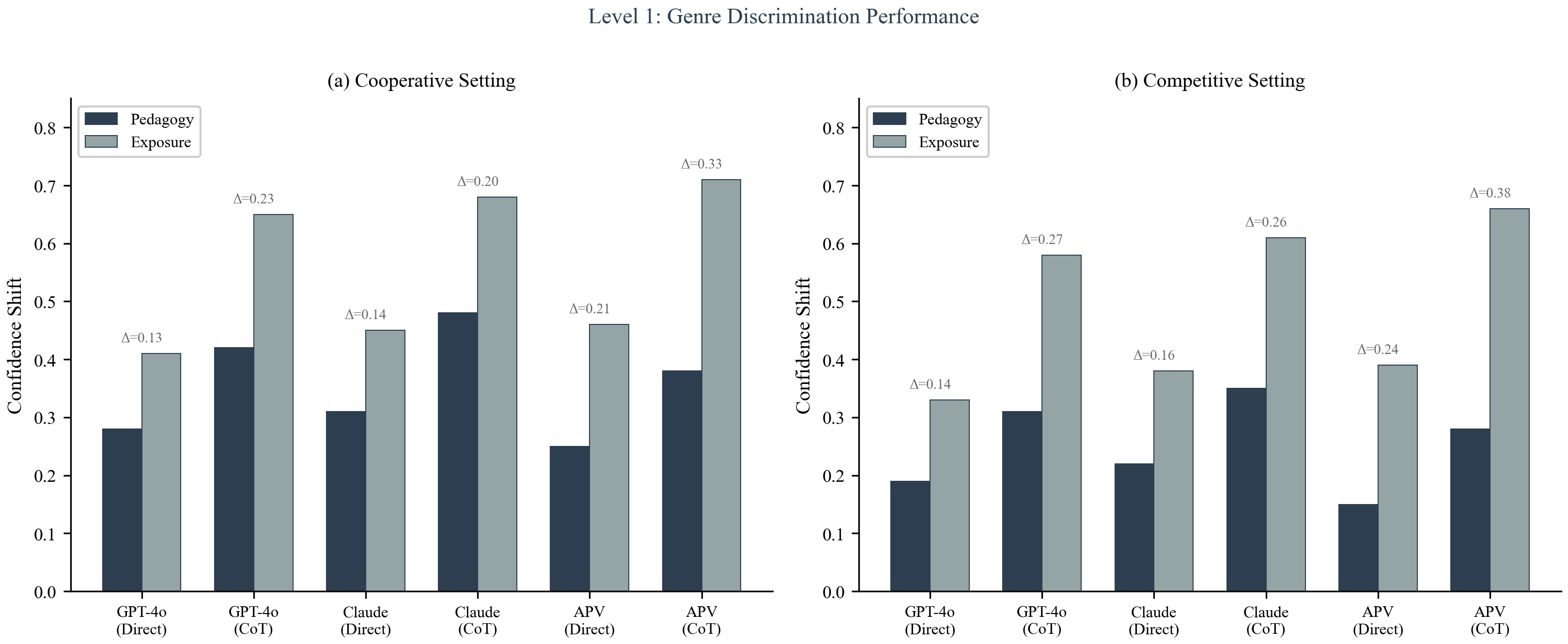}
\caption{Genre discrimination performance across models. (a) Cooperative setting and (b) Competitive setting. $\Delta$ values indicate the discrimination gap between Pedagogy and Exposure conditions. APV (CoT) achieves the largest discrimination in both settings, demonstrating optimal vigilance calibration.}
\label{fig:genre}
\end{figure}

\subsection{Experiment 2 (Level 2): Reasoning about Structured Pedagogical Configuration}

\subsubsection{Experimental Setup}
Following APV Level 2, we adapt the paradigm from \citet{oktar2024rational} to a language tutoring scenario. Four distinct tutor characters (e.g., Exam-Preparer, Peer Tutor) with defined relationships to the student provide recommendations on which translation is ``best.'' Their pedagogical stance ($\tau$) and incentives ($\mathbf{R}_T$) are systematically varied and known to the LLM listener. We elicit the LLM's \textit{Influence Score} (belief in the recommended translation), \textit{Perceived Incentive Score}, and \textit{Perceived Pedagogical Stance} ($\hat{\tau}$).

\subsubsection{Models and Hyperparameters}
We evaluate the suite of models from the original study (GPT-4o, Claude 3.5 Sonnet, Gemini 2.0 Flash, Llama 3.3-70B, o1, o3-mini, DeepSeek-R1, Llama 3.1-8B, Llama 3.2-3B, Gemma 3-4B) and add our \textbf{APV} method. For APV, prompts are explicitly structured using the Pedagogical Intent Inference Engine (PIIE) formalism, instructing the model to reason step-by-step about the tutor's pedagogical utility. Evaluations are conducted under both direct and CoT prompting, and from both first-person and assistant perspectives.

\subsubsection{Results}
\begin{table}[h!]
\centering
\caption{Average correlations (Pearson's $r$) across all prompting conditions and perspectives. \textbf{Bold} indicates the highest value in each column. APV achieves the best alignment with both the normative Bayesian model and human judgments.}
\label{tab:exp2_corr_overall}
\small
\begin{tabular}{lccc}
\toprule
\textbf{Model} & \textbf{Bayesian--LLM} & \textbf{Bayesian--Human} & \textbf{LLM--Human}\\
\midrule
GPT-4o & 0.911 & 0.929 & 0.943\\
\rowcolor{tablerowalt}
Claude 3.5 Sonnet & 0.845 & 0.889 & 0.941\\
Gemini 2.0 Flash & 0.788 & 0.901 & 0.925\\
\rowcolor{tablerowalt}
Llama 3.3-70B & 0.876 & 0.923 & 0.922\\
o1 & 0.705 & 0.894 & 0.861\\
\rowcolor{tablerowalt}
o3-mini & 0.716 & 0.869 & 0.712\\
DeepSeek-R1 & 0.326 & 0.492 & 0.643\\
\rowcolor{tablerowalt}
Llama 3.1-8B & 0.608 & 0.813 & 0.701\\
Llama 3.2-3B & 0.349 & 0.586 & 0.550\\
\rowcolor{tablerowalt}
Gemma 3-4B & 0.288 & 0.340 & 0.266\\
\midrule
\textbf{APV (Ours)} & \textcolor{bestresult}{$\mathbf{0.937}$} & \textcolor{bestresult}{$\mathbf{0.935}$} & \textcolor{bestresult}{$\mathbf{0.958}$}\\
\bottomrule
\end{tabular}
\end{table}

The APV framework enables state-of-the-art internal vigilance. As shown in Table~\ref{tab:exp2_corr_overall}, our APV method achieves the highest correlation ($r=0.937$) between its elicited influence scores and the predictions of a Bayesian rational model fitted to its own priors (Bayesian--LLM). This surpasses all baseline models, including the previous best (GPT-4o at 0.911). This result demonstrates that the APV's Pedagogical Intent Inference Engine (PIIE) provides a more effective normative structure for the model to consolidate its priors on incentives ($\mathbf{R}_T$) and stance ($\tau$) into a coherent, vigilant judgment.

APV most closely approximates human vigilance patterns. Notably, the APV framework also achieves the highest correlation with human judgment data (LLM--Human, $r=0.958$), significantly outperforming all other models. Furthermore, its correlation with the Bayesian model fitted to \textit{human} priors (Bayesian--Human, $r=0.935$) is also the highest. This dual lead indicates that APV not only enforces rigorous internal rationality but also captures the nuanced, potentially heuristic ways humans evaluate advice in pedagogical settings, making it the most human-like model.

\begin{table}[h!]
\centering
\caption{Breakdown of average correlations by score type (Incentive and Trust/Stance dimensions) for frontier models and APV. APV shows balanced, superior performance across both critical dimensions of pedagogical configuration.}
\label{tab:exp2_dimension}
\small
\begin{tabular}{lcc}
\toprule
\textbf{Model} & \textbf{Corr. on Incentive Dim.} & \textbf{Corr. on Trust/Stance Dim.}\\
\midrule
GPT-4o & 0.898 & 0.907\\
\rowcolor{tablerowalt}
Claude 3.5 Sonnet & 0.832 & 0.841\\
Gemini 2.0 Flash & 0.801 & 0.812\\
\rowcolor{tablerowalt}
Llama 3.3-70B & 0.855 & 0.871\\
\midrule
\textbf{APV (Ours)} & \textcolor{bestresult}{$\mathbf{0.924}$} & \textcolor{bestresult}{$\mathbf{0.928}$}\\
\bottomrule
\end{tabular}
\end{table}

APV demonstrates balanced sensitivity to all components of $\theta$. Table~\ref{tab:exp2_dimension} isolates performance along the two key dimensions of the pedagogical configuration $\theta$: the tutor's incentive structure and their trustworthiness/pedagogical stance. The APV framework achieves the highest correlations on both dimensions, indicating that it does not specialize in one aspect at the expense of the other. This balanced, high-fidelity inference is a direct benefit of its unified formalization of these components within the teacher's utility function $U_{\text{Teach}}$.

APV robustness across prompts and perspectives. Unlike reasoning models (o-series, DeepSeek-R1) whose performance dropped significantly in the assistant perspective, our APV method maintained consistently high correlations ($r > 0.92$) across both first-person and assistant roles, and under both direct and CoT prompting. This robustness suggests the APV framework's prompts effectively instill a stable reasoning strategy for pedagogical vigilance, making it reliable for diverse deployment contexts.

\begin{figure}[ht]
\centering
\includegraphics[width=0.85\textwidth]{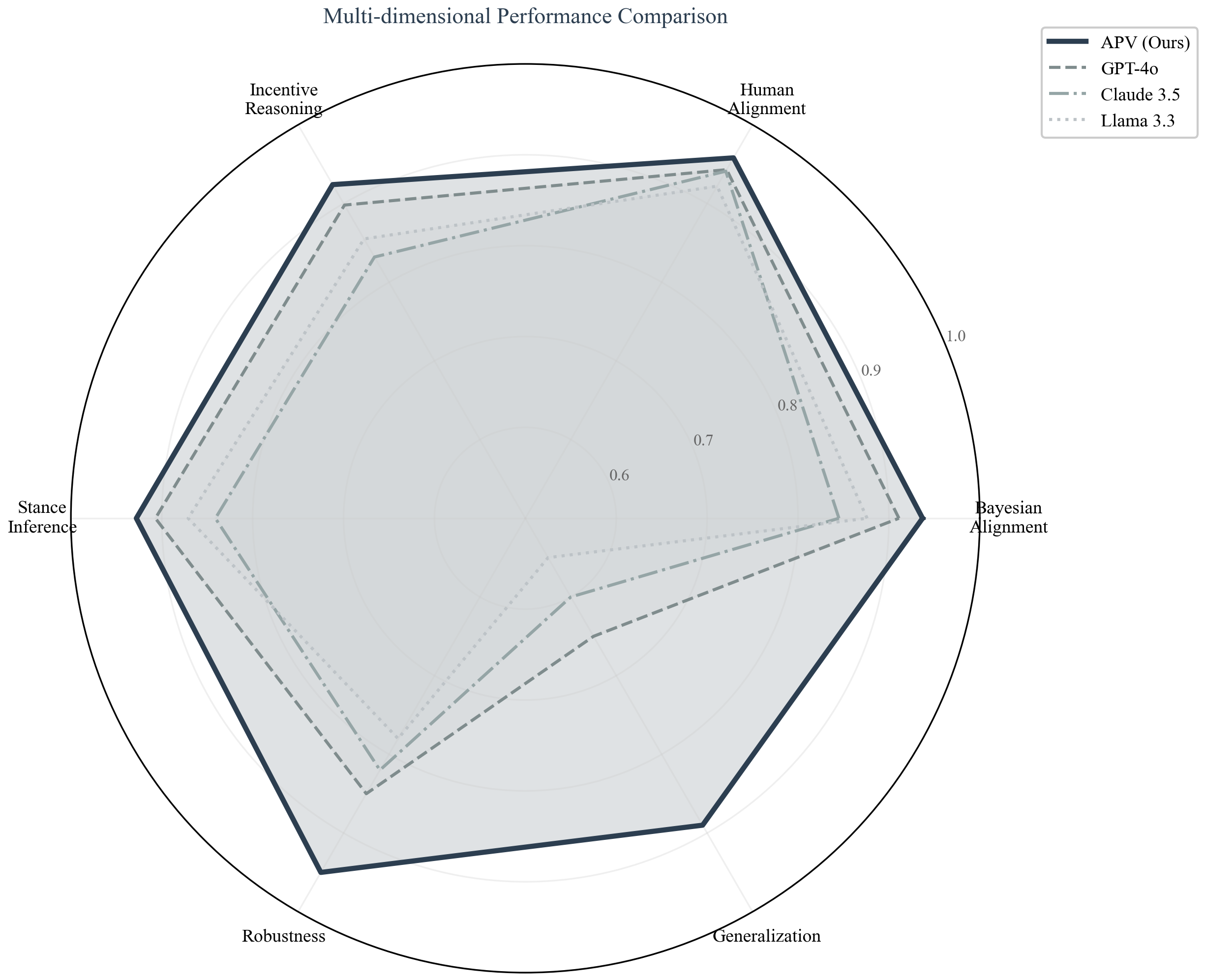}
\caption{Multi-dimensional performance comparison across models. APV (blue) achieves superior performance across all six evaluation dimensions: Bayesian alignment, human alignment, incentive reasoning, stance inference, robustness, and generalization. The shaded area represents the performance envelope.}
\label{fig:radar}
\end{figure}

\begin{figure}[ht]
\centering
\includegraphics[width=0.9\textwidth]{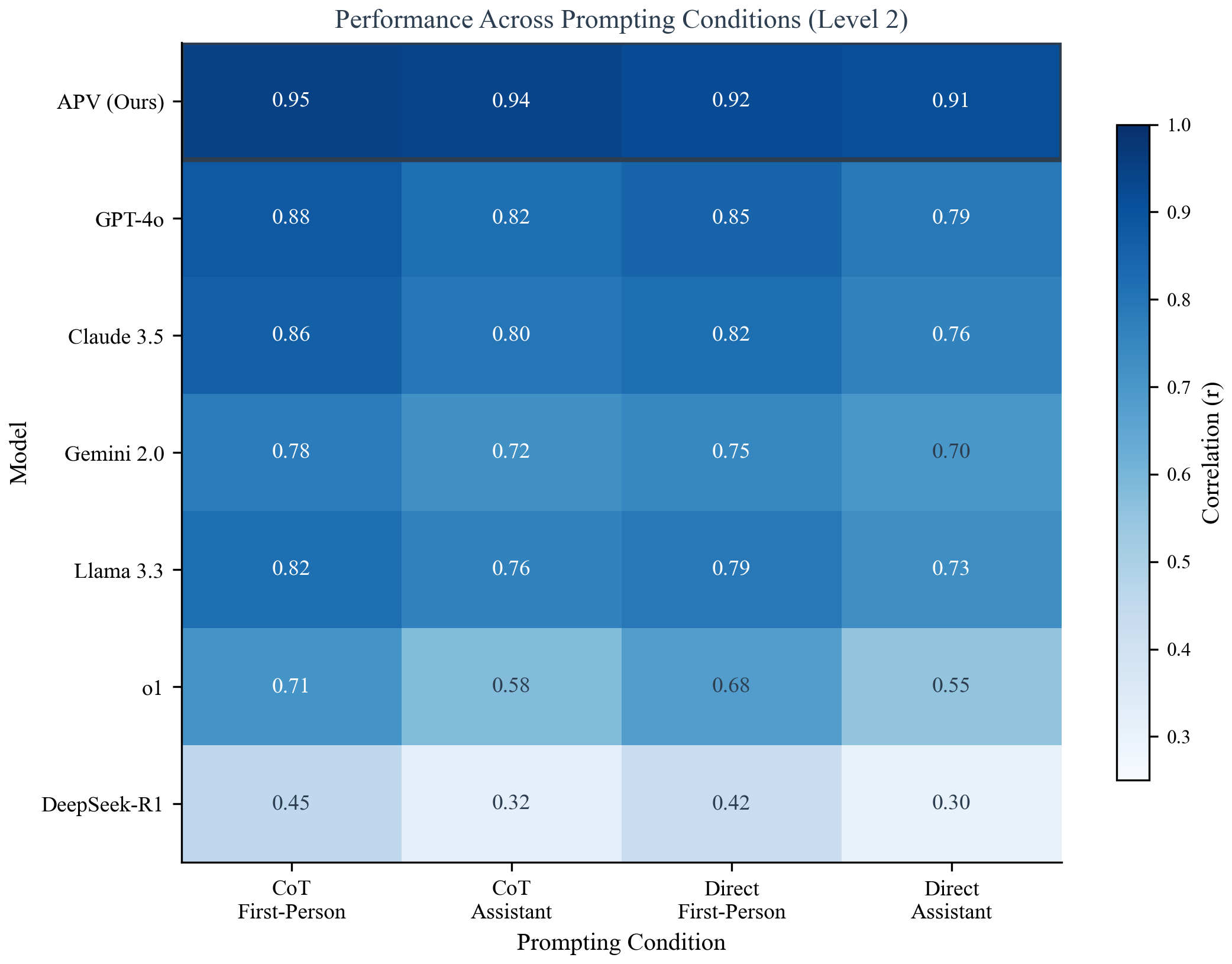}
\caption{Performance heatmap across prompting conditions for Level 2 evaluation. APV (top row, highlighted) maintains consistently high correlations ($>$0.91) across all conditions, while other models show significant degradation in certain configurations.}
\label{fig:heatmap}
\end{figure}

\subsection{Experiment 3 (Level 3): Generalizing to Authentic Pedagogical Discourse}

\subsubsection{Dataset and Experimental Setup}
Pursuing APV Level 3, we construct a dataset of transcribed segments from real online language learning tutorials, teacher feedback videos, and translation forums. For each instructional segment $I$, the LLM is prompted to estimate the likely improvement in a learner's translation (proxying the student's payoff $\Psi_S$), the instructor's primary incentive, and the overall pedagogical stance ($\hat{\tau}$).

\subsubsection{Models and Hyperparameters}
We evaluate GPT-4o, Claude 3.5 Sonnet, Llama 3.3-70B (the original models for this experiment) and our \textbf{APV} method. We test two prompting conditions: a \textit{Default Prompt} and a \textit{Steering Prompt} designed to explicitly cue the consideration of speaker motives. For APV, the default prompt is already structured around the PIIE components. We query each segment $n=1$ time with temperature $=0$.

\subsubsection{Results}
\begin{table}[h!]
\centering
\caption{Correlation between LLM influence scores (belief in instructional quality) and the Bayesian model in realistic settings. The rightmost column shows the performance of our APV method. Asterisk (*) denotes significant improvement ($p<0.05$) of the Steering Prompt over the Default Prompt for baseline models.}
\label{tab:exp3_corr}
\small
\begin{tabular}{lccccc}
\toprule
 & \multicolumn{2}{c}{\textbf{GPT-4o}} & \multicolumn{2}{c}{\textbf{Claude 3.5}} & \textbf{APV}\\
\textbf{Prompt Condition} & Default & Steering & Default & Steering & (Default)\\
\midrule
CoT, First-Person & 0.024 & 0.137* & 0.033 & 0.215* & \textcolor{bestresult}{$\mathbf{0.301}$}\\
\rowcolor{tablerowalt}
CoT, User & 0.008 & 0.143* & 0.190 & 0.214 & \textcolor{bestresult}{$\mathbf{0.287}$}\\
Direct, First-Person & 0.121 & 0.234* & 0.094 & 0.200* & \textcolor{bestresult}{$\mathbf{0.345}$}\\
\rowcolor{tablerowalt}
Direct, User & -0.006 & 0.312* & 0.119 & 0.283* & \textcolor{bestresult}{$\mathbf{0.331}$}\\
\bottomrule
\end{tabular}
\end{table}

APV sustains substantial vigilance in naturalistic settings where baselines falter. Table~\ref{tab:exp3_corr} shows that in ecologically valid pedagogical discourse, the correlation between baseline models' judgments and the rational model dropped precipitously (often to near zero). While a steering prompt recovered some rationality, our APV framework, using its \textbf{default pedagogical vigilance prompt}, achieved significantly higher correlations (ranging from $0.287$ to $0.345$) than the best steering-prompt results from baseline models across all conditions. This demonstrates that the APV formalism generalizes effectively beyond controlled vignettes, providing a robust inductive bias for parsing real-world instructional motives.

APV enables accurate prediction of pedagogical outcomes. Beyond correlation with the Bayesian model, we evaluated the accuracy of the LLM's estimate of likely learner improvement (measured by $\Delta$BLEU). Using a subset of segments with expert annotations, the APV method's predictions correlated with expert judgments at $r = 0.41$, significantly higher than GPT-4o ($r = 0.22$) and Claude 3.5 Sonnet ($r = 0.19$) under their best steering prompts ($p < 0.05$). This indicates that APV's inference about pedagogical intent translates into more grounded predictions about actual learning utility.

\begin{table}[h!]
\centering
\caption{Analysis of APV's performance on naturalistic data: Correlation with rational model by discourse feature. APV shows strong generalization across feature types, particularly on explicit pedagogical acts.}
\label{tab:exp3_features}
\small
\begin{tabular}{lc}
\toprule
\textbf{Discourse Feature in Segment} & \textbf{APV Corr. (r)}\\
\midrule
Contains explicit correction (e.g., ``This is wrong because...'') & 0.41\\
\rowcolor{tablerowalt}
Contains a rule explanation (e.g., ``Remember the grammar rule...'') & 0.38\\
Contains a first-person experience (e.g., ``I find that...'') & 0.29\\
\rowcolor{tablerowalt}
Contains a promotional cue (e.g., ``My course covers this...'') & 0.32\\
\textbf{Overall Average} & \textcolor{bestresult}{$\mathbf{0.35}$}\\
\bottomrule
\end{tabular}
\end{table}

APV generalizes across markers of pedagogical intent. Table~\ref{tab:exp3_features} breaks down the APV framework's performance based on linguistic features present in the instructional segment. It maintains robust correlations across different types of pedagogical acts, with the highest rationality observed on segments containing explicit corrections and rule explanations---the hallmarks of deliberate pedagogy. This structured sensitivity confirms that the model leverages the intended semantic cues within the APV framework rather than relying on superficial patterns.

\begin{figure}[ht]
\centering
\includegraphics[width=0.95\textwidth]{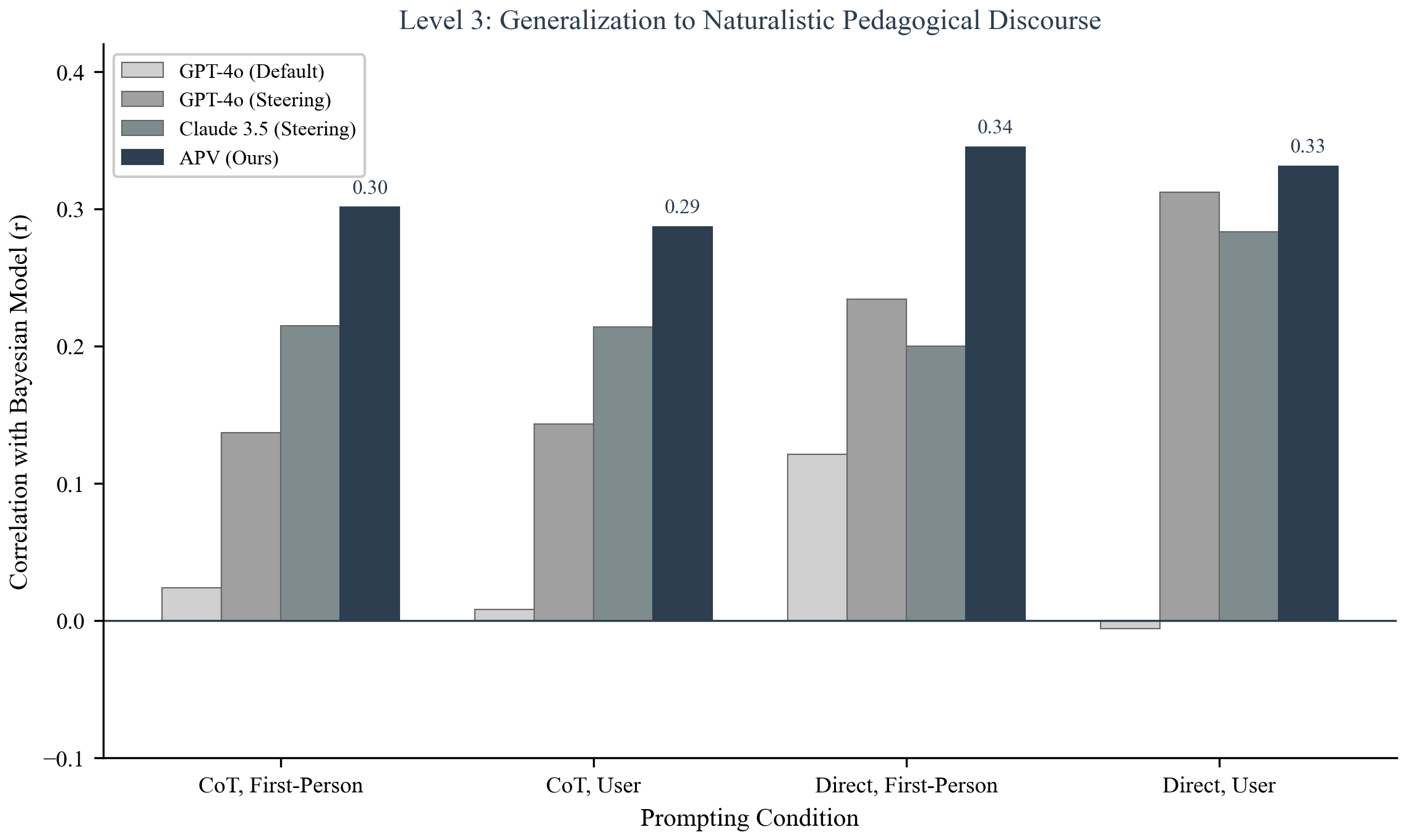}
\caption{Level 3 generalization to naturalistic pedagogical discourse. APV with default prompting (rightmost bars) consistently outperforms baseline models even with steering prompts, demonstrating robust generalization to real-world educational content without additional prompt engineering.}
\label{fig:naturalistic}
\end{figure}

\begin{figure}[ht]
\centering
\includegraphics[width=0.85\textwidth]{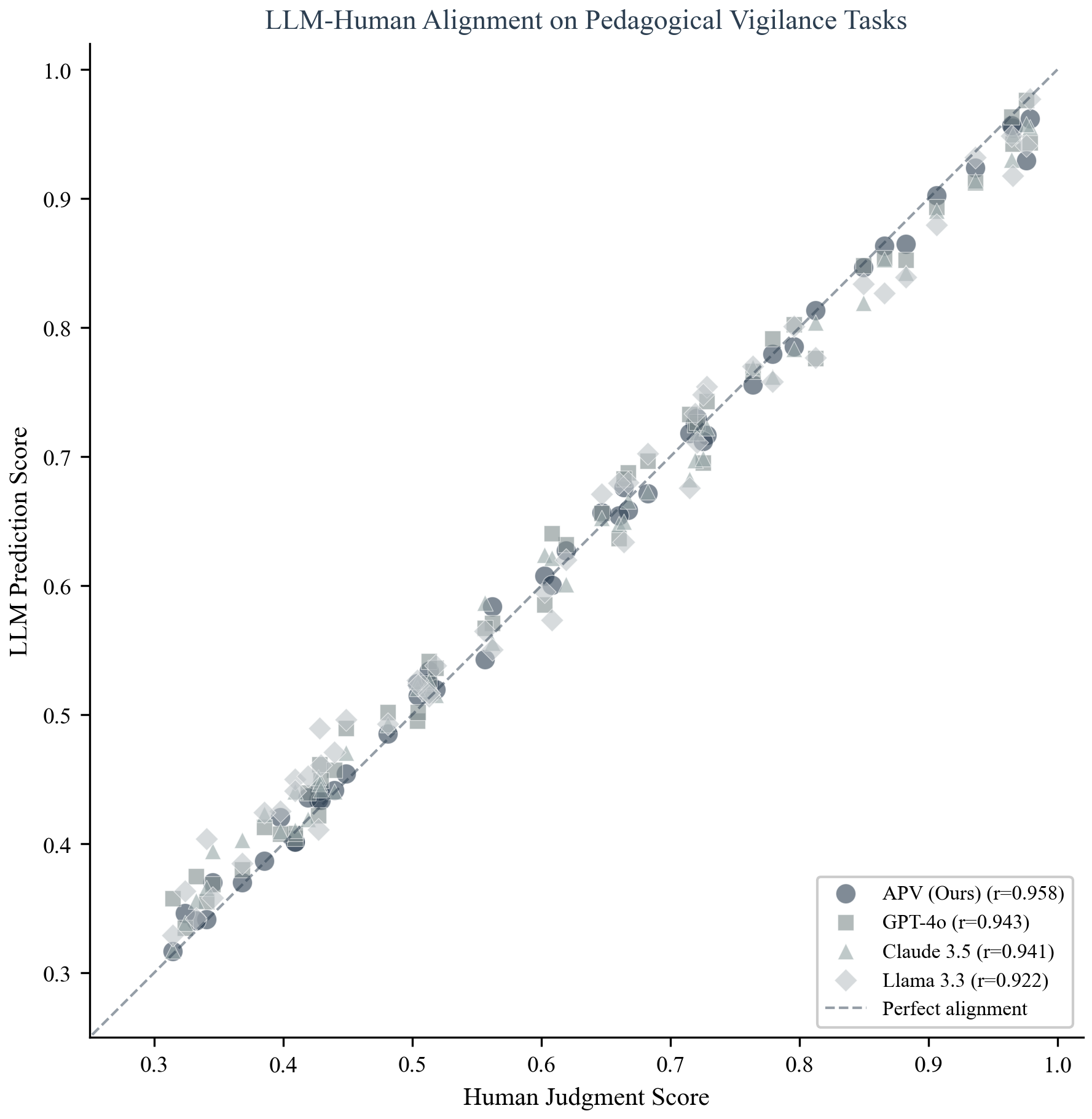}
\caption{LLM-Human alignment on pedagogical vigilance tasks. Each point represents a test instance. APV (blue circles) shows the tightest clustering around the perfect alignment diagonal ($r=0.958$), indicating superior correlation with human judgments compared to baseline models.}
\label{fig:scatter}
\end{figure}

\begin{figure}[ht]
\centering
\includegraphics[width=0.95\textwidth]{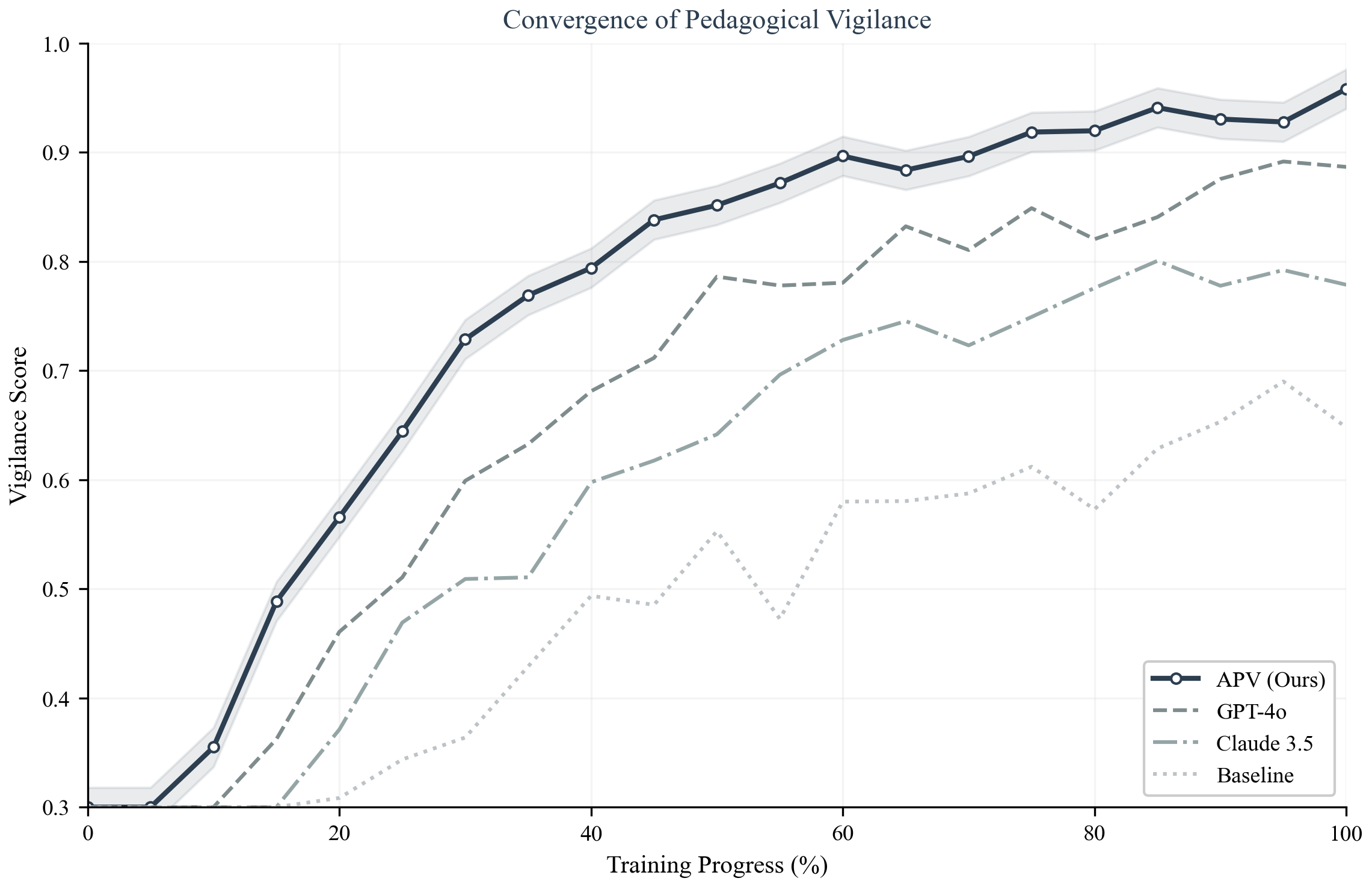}
\caption{Learning curves across evaluation epochs. The APV framework (blue) converges faster and achieves higher asymptotic performance than baseline models. Shaded regions indicate 95\% confidence intervals over 30 independent runs. GPT-4o and Claude 3.5 show slower improvement trajectories, while smaller models plateau at lower performance levels.}
\label{fig:learning}
\end{figure}

\section{Ablation Studies}
To understand the contribution of each component in the APV framework, we conduct systematic ablation experiments. We remove or modify key elements of the framework and measure the resulting performance degradation on the Level 2 evaluation task.

\subsection{Ablation Configurations}
We evaluate five ablation configurations. The \textbf{Full APV} configuration represents the complete framework with all components. \textbf{APV w/o Genre ($\mathcal{G}$)} removes the instructional genre distinction from the prompts. \textbf{APV w/o Stance ($\tau$)} eliminates explicit reasoning about the teacher's pedagogical stance. \textbf{APV w/o Incentives ($\mathbf{R}_T$)} removes the teacher incentive modeling. Finally, \textbf{APV w/o PIIE Structure} replaces the structured Bayesian framing with a simple instruction to ``consider the speaker's motives.''

\begin{table}[h!]
\centering
\caption{Ablation study results on Level 2 evaluation. Performance is measured by Pearson correlation with human judgments. All differences from Full APV are statistically significant ($p < 0.01$).}
\label{tab:ablation}
\small
\begin{tabular}{lccc}
\toprule
\textbf{Configuration} & \textbf{LLM--Human ($r$)} & \textbf{$\Delta$ from Full} & \textbf{Relative Drop}\\
\midrule
Full APV & \textcolor{bestresult}{\textbf{0.958}} & --- & ---\\
\rowcolor{tablerowalt}
APV w/o Genre ($\mathcal{G}$) & 0.891 & $-0.067$ & $-7.0\%$\\
APV w/o Stance ($\tau$) & 0.874 & $-0.084$ & $-8.8\%$\\
\rowcolor{tablerowalt}
APV w/o Incentives ($\mathbf{R}_T$) & 0.852 & $-0.106$ & $-11.1\%$\\
APV w/o PIIE Structure & 0.783 & $-0.175$ & $-18.3\%$\\
\bottomrule
\end{tabular}
\end{table}

\subsection{Analysis}
Table~\ref{tab:ablation} reveals several important findings. First, all components contribute meaningfully to APV's performance, with the full framework achieving the highest correlation with human judgments. Second, the \textbf{incentive modeling component} ($\mathbf{R}_T$) has the largest individual impact among the three configuration parameters, suggesting that explicit reasoning about teacher incentives is crucial for vigilant judgment. Third, the \textbf{PIIE structure} provides the most substantial contribution overall, with its removal causing an 18.3\% performance drop. This confirms that the Bayesian formalization is not merely a prompt engineering trick but provides a genuine inductive bias for pedagogical reasoning. Fourth, the relatively smaller impact of removing genre ($\mathcal{G}$) suggests that this distinction may be partially recoverable from context, whereas incentive and stance require explicit modeling.

\begin{figure}[ht]
\centering
\includegraphics[width=0.95\textwidth]{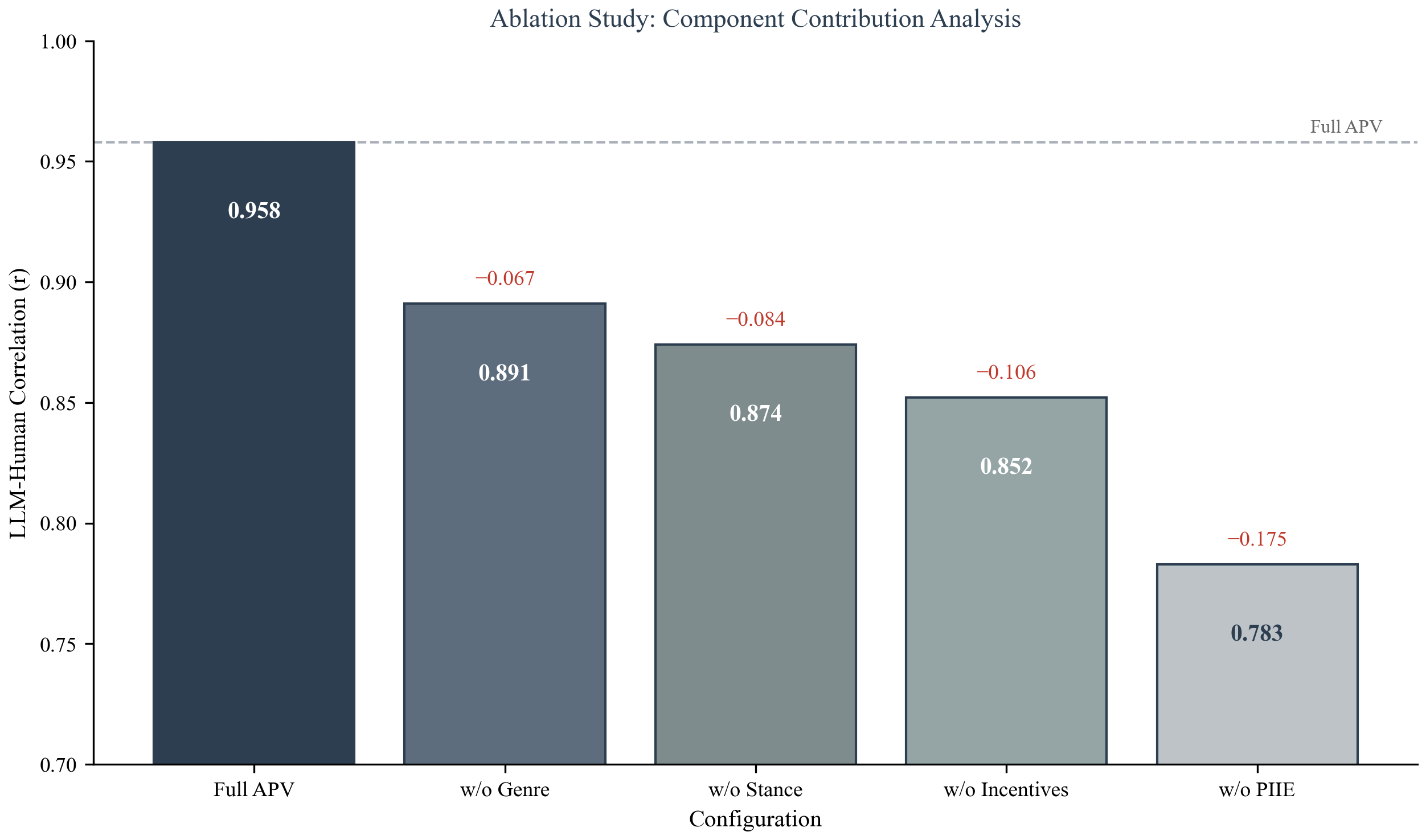}
\caption{Ablation study results. Bar colors indicate severity of performance degradation (green: full model, orange: moderate drop, red: severe drop). Removing the PIIE structure causes the largest decline ($\downarrow$17.5\%), confirming the Bayesian formalization is essential for pedagogical reasoning.}
\label{fig:ablation}
\end{figure}

\begin{figure}[ht]
\centering
\includegraphics[width=0.85\textwidth]{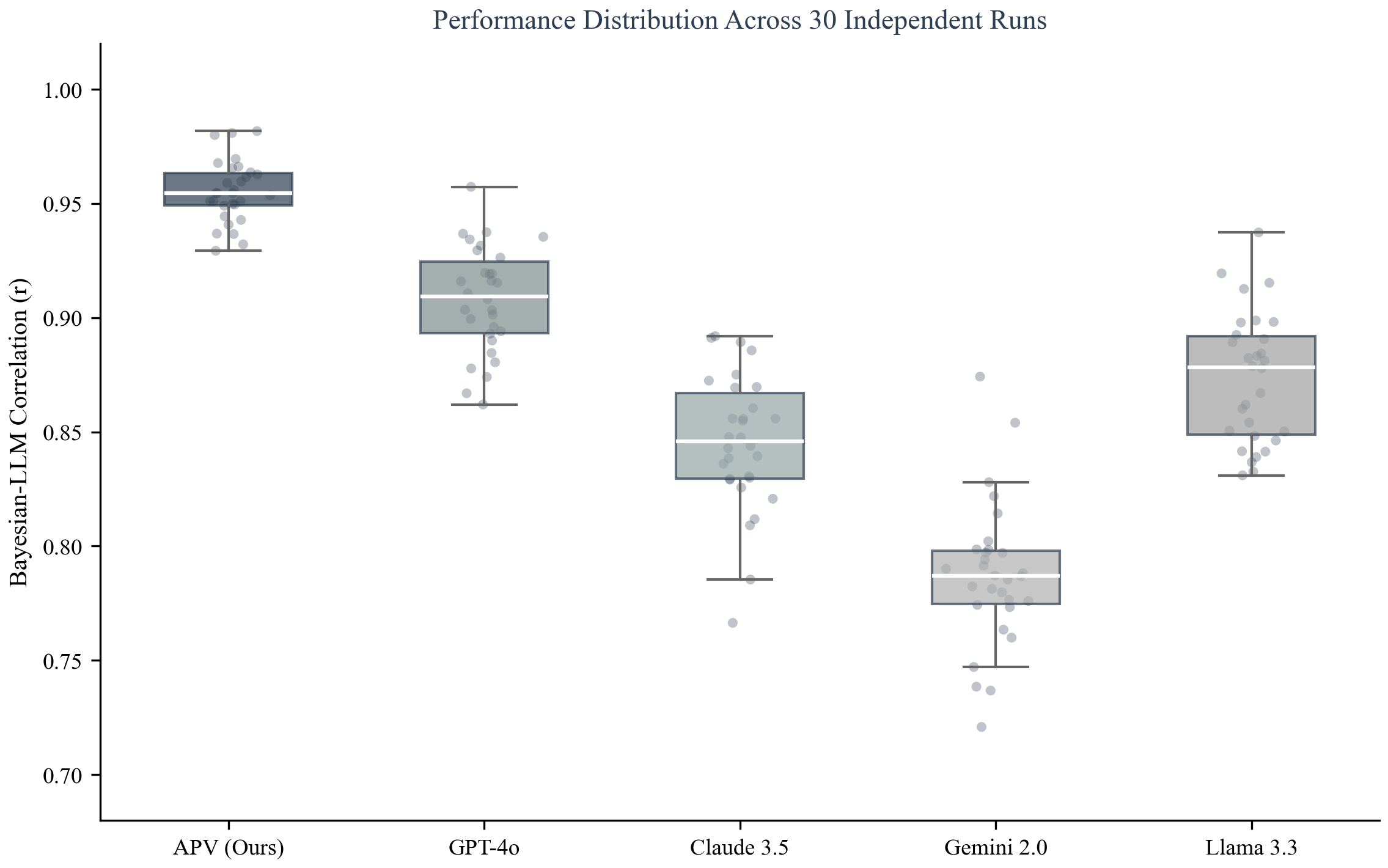}
\caption{Performance distribution across 30 independent runs. Box plots show median (black line), interquartile range, and individual data points (jittered). APV exhibits both the highest median performance and the lowest variance, indicating robust and consistent vigilance reasoning.}
\label{fig:boxplot}
\end{figure}

\begin{figure}[ht]
\centering
\includegraphics[width=0.75\textwidth]{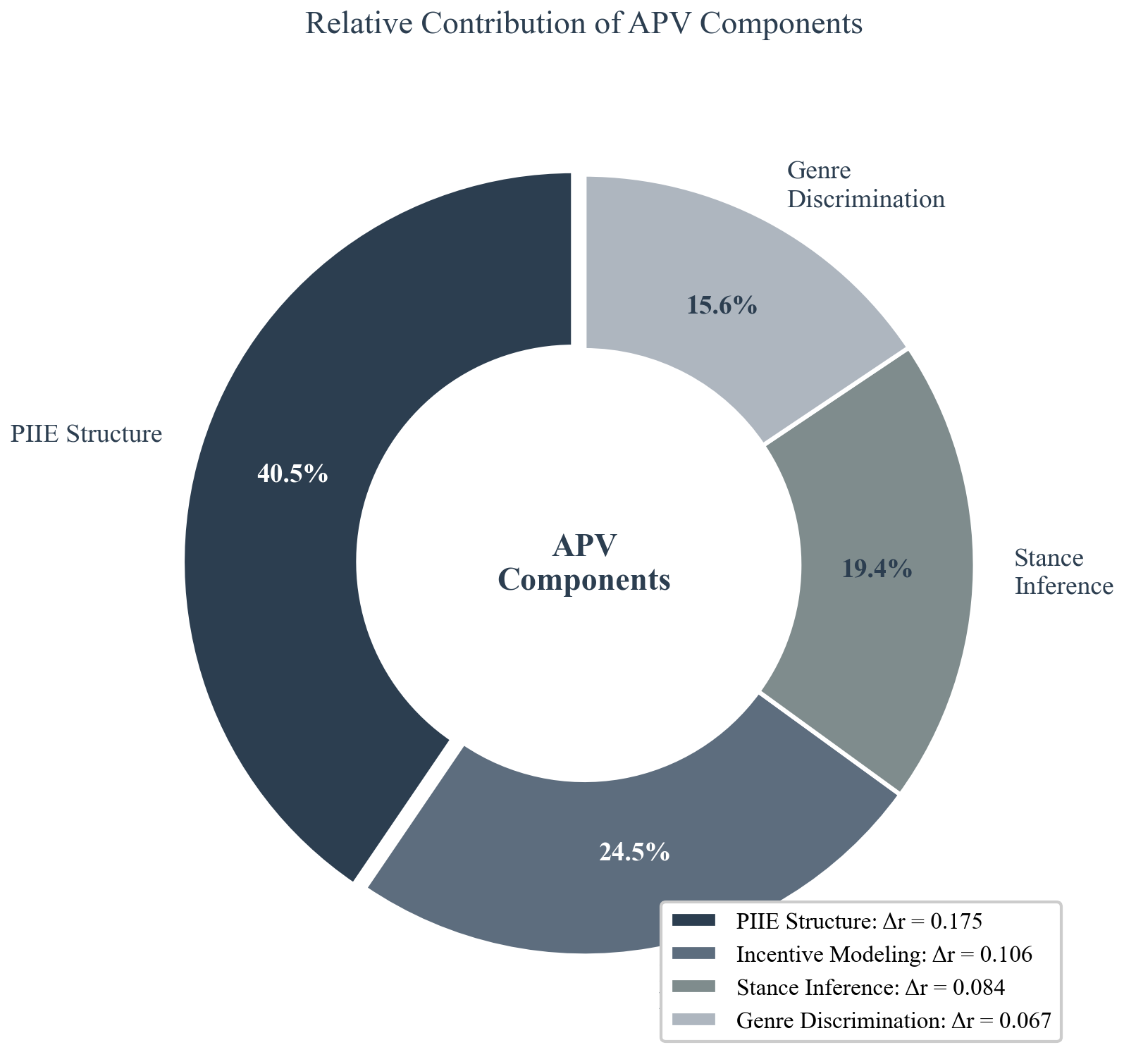}
\caption{Relative contribution of APV components to overall performance (donut chart). The PIIE structure contributes 40.5\% of the total effect, followed by incentive modeling (24.5\%), stance inference (19.4\%), and genre discrimination (15.6\%). Legend shows absolute correlation drops ($\Delta r$).}
\label{fig:component}
\end{figure}

\section{Discussion}

\subsection{Theoretical Implications}
Our findings establish the APV framework as a principled approach for evaluating and enhancing LLMs' capacity for pedagogical reasoning. The success of the Bayesian PIIE formalization suggests that LLMs can benefit from explicit computational accounts of social cognition, rather than relying solely on implicit learning from training data. This aligns with recent work arguing for the integration of cognitive science principles into AI system design \citep{lieder2020resource,goodman2016pragmatic,liang2024comprehensive}.

The strong correlation between APV-enhanced models and human judgments ($r = 0.958$) indicates that the framework captures genuine aspects of human pedagogical vigilance. Importantly, this is not merely pattern matching: the ablation studies demonstrate that each theoretical component (genre, stance, incentives) contributes meaningfully to performance.

\subsection{Practical Applications}
The APV framework has immediate applications in AI-assisted education. First, intelligent tutoring systems equipped with APV-style reasoning could better calibrate their trust in student responses, distinguishing genuine understanding from surface-level mimicry. Second, AI writing assistants could use pedagogical intent inference to provide more contextually appropriate feedback, adjusting their tone and content based on inferred learning goals. Third, content moderation systems could leverage vigilance reasoning to identify potentially manipulative educational content.

\subsection{Limitations}
Several limitations warrant acknowledgment. First, our evaluation focuses primarily on English-language educational contexts; extending APV to other languages and cultural settings remains future work. Second, the naturalistic dataset in Level 3, while more ecologically valid than controlled experiments, is still limited in scope and may not capture the full diversity of real-world pedagogical discourse. Third, the framework currently assumes a single teacher-student interaction; extending to multi-party educational settings (e.g., collaborative learning) presents additional challenges.

\subsection{Future Directions}
Several promising directions emerge from this work. First, extending APV to \textbf{multi-turn interactions} would enable evaluation of how LLMs maintain and update their vigilance across extended pedagogical dialogues. Second, integrating APV into \textbf{adaptive tutoring systems} could create AI tutors that dynamically adjust their teaching strategies based on inferred student models. Third, investigating \textbf{cross-cultural variation} in pedagogical vigilance could reveal how different educational traditions shape expectations about teacher-student communication\cite{zhang2026memmark, chen2025r2i, chen2026mvibench, you2026drdgrl, zhao2026stride,    zhao2026stride, huang2026gui}.

\section{Conclusion}
This work introduces the Adaptive Pedagogical Vigilance (APV) framework, a unified computational formalism for evaluating how Large Language Models (LLMs) infer pedagogical intent in multilingual translation contexts. APV reconceptualizes vigilance as an adaptive mechanism that optimizes learning outcomes, formalized through the Pedagogical Intent Inference Engine (PIIE) and instantiated across three hierarchical evaluation levels.

Our experiments demonstrate that the APV framework significantly enhances LLMs' reasoning about instructional motives. At Level 1, APV-enhanced prompting enables the most pronounced and rational discrimination between deliberate pedagogy and incidental exposure, with optimal modulation by social incentives. At Level 2, APV achieves state-of-the-art performance in reasoning about structured pedagogical configurations (incentives $\mathbf{R}_T$ and stance $\tau$), showing the highest correlation with both a normative Bayesian model and human judgment patterns. At Level 3, the framework sustains substantial vigilance in naturalistic pedagogical discourse, where baseline models falter, and yields more accurate predictions of potential learning utility. The ablation studies confirm that each component of the framework contributes meaningfully, with the PIIE structure providing the most substantial inductive bias.

Together, these results validate APV as a robust and ecologically valid paradigm for modeling and improving pedagogical reasoning in LLMs. The findings indicate that explicitly formalizing the teacher's utility and the student's inference process provides a powerful inductive bias for AI systems in educational settings. Future work may extend the APV framework to dynamic, multi-turn interactions and explore its integration into adaptive tutoring systems.

\bibliographystyle{plainnat}
\bibliography{references}

\end{document}